%% file: main.tex
\documentclass[10pt,twocolumn,letterpaper]{article}

\usepackage[pagenumbers]{template/cvpr} 

\usepackage[table]{xcolor}
\usepackage{epsfig}
\usepackage{graphicx, amsmath, amssymb, subcaption, multirow, overpic, textpos, pifont, xfrac}
\usepackage{floatrow}
\usepackage{caption}

\usepackage[pagebackref,breaklinks,colorlinks]{hyperref}

\input{definitions}

\usepackage[capitalize]{cleveref}
\crefname{section}{Sec.}{Secs.}
\Crefname{section}{Section}{Sections}
\Crefname{table}{Table}{Tables}
\crefname{table}{Tab.}{Tabs.}

\def\cvprPaperID{no} 
\def\confName{CVPR}
\def\confYear{2023}

\begin{document}

\title{Token Merging for Fast Stable Diffusion}

\author{Daniel Bolya $\quad$ Judy Hoffman\\
Georgia Tech\\
{\tt\small \{dbolya,judy\}@gatech.edu}
}
\maketitle

\begin{abstract}
The landscape of image generation has been forever changed by open vocabulary diffusion models. However, at their core these models use transformers, which makes generation slow. Better implementations to increase the throughput of these transformers have emerged, but they still evaluate the entire model. In this paper, we instead speed up diffusion models by exploiting natural redundancy in generated images by merging redundant tokens. After making some diffusion-specific improvements to Token Merging (ToMe), our ToMe for Stable Diffusion can reduce the number of tokens in an existing Stable Diffusion model by up to 60\% while still producing high quality images without any extra training. In the process, we speed up image generation by up to $2\times$ and reduce memory consumption by up to $5.6\times$. Furthermore, this speed-up stacks with efficient implementations such as xFormers, minimally impacting quality while being up to $5.4\times$ faster for large images. Code is available at \href{https://github.com/dbolya/tomesd}{https://github.com/dbolya/tomesd}.
\end{abstract}

\section{Introduction} \label{sec:intro}
    With the rise of powerful diffusion \cite{diffusion,guided-diffusion} models such as DALL-E 2 \cite{dalle2}, Imagen \cite{imagen}, and Stable Diffusion \cite{ldm}, generating high quality images has never been easier. However, running these models can be expensive, especially for large images. All of these methods function by denoising images through several evaluations of a transformer \cite{attnisallyouneed} backbone, meaning that computation scales with the \textit{square} of the number of tokens (and thus also the square of pixels).

    Several existing methods to speed up transformers have already been successfully applied to open-source diffusion models such as Stable Diffusion. Flash Attention \cite{flashattn} computes attention efficiently by cleverly accounting for memory bandwidth. XFormers \cite{xformers} contains several optimized implementation of transformer components. And as of PyTorch 2.0, these optimizations are natively available \cite{pytorch}.

    However, none of these approaches \textit{reduce} the amount of work necessary---they still evaluate the transformer on every token.
    Most images, including those generated by diffusion models, have a high amount of \textit{redundancy}. And thus, performing computation on \textit{every token} is a waste of resources.
    Recent work in token reduction such as token pruning \cite{dynamicvit,avit,adavit,spvit} and token merging \cite{tokenpooling,tokenlearner,tome} have shown the ability to \textit{remove} these redundant tokens in transformers to speed up evaluation with a small accuracy drop.

    Though most of these methods require re-training the model (which would be prohibitively expensive for e.g., Stable Diffusion), Token Merging (ToMe) \cite{tome} stands out in particular by \textit{not requiring any additional training}. While the authors only apply it to ViT \cite{vit} for classification, they claim that it should also work for downstream tasks.

\begin{figure}[t!]
    \centering
    \includegraphics[width=1\linewidth]{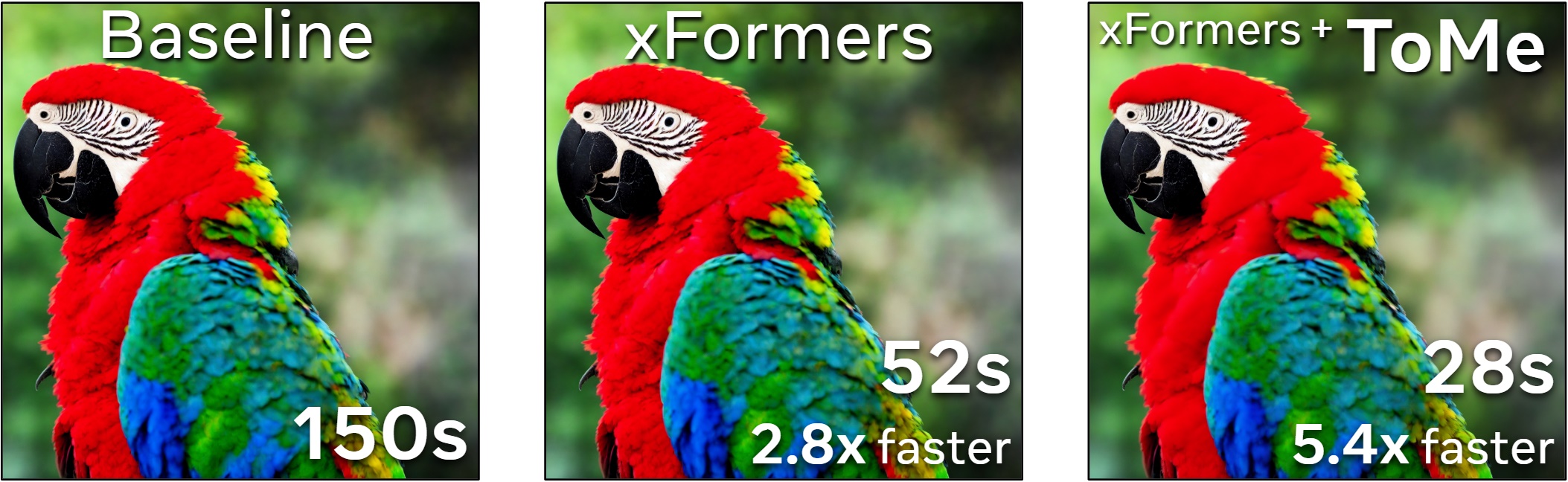}
    \vspace{-1.8em}
    \caption{{\bf Token Merging for Stable Diffusion.} When applied properly, ToMe \cite{tome} can significantly increase the speed of image generation without jeopardizing quality. Moreover, its benefits \textit{stack} with existing methods such as xFormers \cite{xformers}. With ToMe and xFormers together, this $2048\times2048$ image generated in just \textit{28 seconds} on a 4090, which is \textbf{5.4$\times$ faster} than the original model.  }
    \label{fig:teaser}
\end{figure}

    In this paper, we put that to the test by applying ToMe to Stable Diffusion. Out of the box, a na\"ive application can speed up diffusion by up to $2\times$ and reduce memory consumption by $4\times$ (Tab.~\ref{tab:naive_tome}), but the resulting image quality suffers greatly (Fig.~\ref{fig:naive_tome_vis}). To address this, we introduce new techniques for token partitioning (Fig.~\ref{fig:destination_set}) and perform several experiments to decide how to apply ToMe (Tab.~\ref{tab:ablations}). As a result, we can keep the speed and improve the memory benefits of ToMe, while producing images extremely close to the original model (Fig.~\ref{fig:results}, Tab.~\ref{tab:results}). Furthermore, this speed-up \textit{stacks} with implementations such as xFormers (Fig.~\ref{fig:teaser}).

\section{Background} \label{sec:background}
In this work, our goal is to speed up an off-the-shelf Stable Diffusion \cite{ldm} model \textit{without training} using ToMe \cite{tome}.

\begin{figure}[t]
    \centering
    \includegraphics[width=1\linewidth]{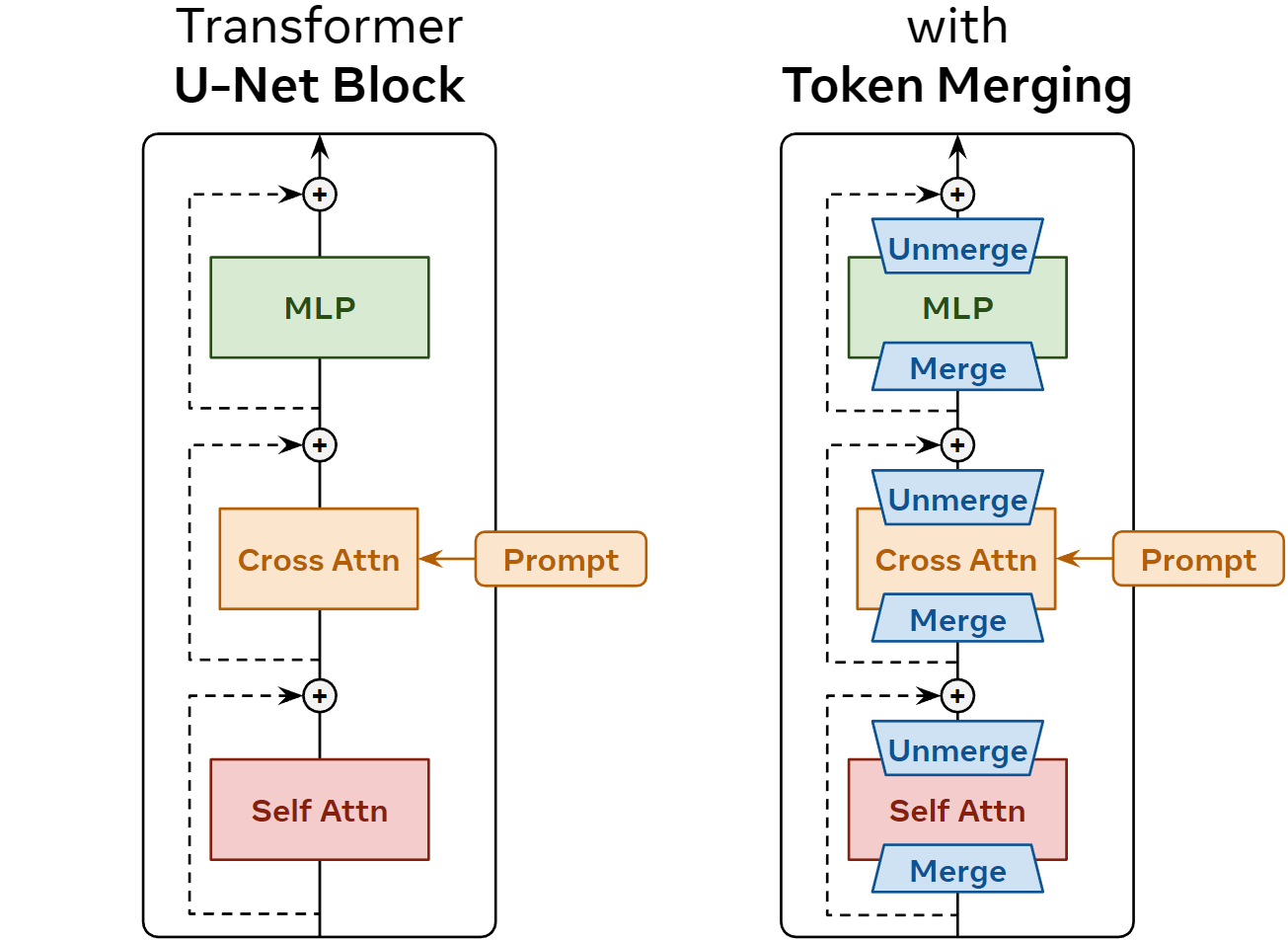}
    \vspace{-1.8em}
    \caption{{\bf A U-Net Block} with ToMe applied. Stable Diffusion \cite{ldm} uses a U-Net \cite{unet} model with transformer-based blocks \cite{attnisallyouneed}. We apply ToMe \cite{tome} by merging tokens before each component of the block and unmerging after to reduce compute costs (Sec.~\ref{sec:approach}). }
    \label{fig:unet_block}
\end{figure}

\paragraph{Stable Diffusion.} Diffusion models \cite{nonequilibrium_thermodynamics,diffusion,guided-diffusion} generate images by repeatedly denoising some initial noise over some number of diffusion steps. Like most modern large diffusion models, Stable Diffusion uses a U-Net \cite{unet} with transformer-based blocks. Thus, it first encodes the current noised image as a set of tokens, then passes it through a series of transformer blocks. Each transformer block has the standard self attention \cite{attnisallyouneed} and multi-layer perception (mlp) modules, with the addition of a cross attention module to condition on the prompt (see Fig.~\ref{fig:unet_block}).

\paragraph{Token Merging.}
Token Merging (ToMe) \cite{tome} reduces the number of tokens in a transformer gradually by merging $r$ tokens in each block. To do this efficiently, it partitions the tokens into a \textcolor{srccolor}{source} (\src{src}) and \textcolor{dstcolor}{destination} (\dst{dst}) set. Then, it merges the $r$ \textit{most similar} tokens from \src{src} \textit{into} \dst{dst}, reducing the number of tokens by $r$, making the next block faster.

\section{Token Merging for Stable Diffusion} \label{sec:approach}
While ToMe as described in Sec.~\ref{sec:background} works well for classification, it's not entirely straightforward to apply it to a dense prediction task like diffusion. While classification only needs a single token to make a prediction, diffusion needs to know the noise to remove for \textit{every token}. Thus, we need to introduce the concept of \textit{unmerging}.

\subsection{Defining Unmerging}
While other token reduction methods such as pruning (e.g., \cite{dynamicvit}) \textit{remove} tokens, ToMe is different in that it \textit{merges} them. And if we have information about what tokens we merged, we have enough information to then \textit{unmerge} those same tokens. This is crucial for a dense prediction task, where we really do need every token.

\begin{figure}[t]
    \centering
    \includegraphics[width=1\linewidth]{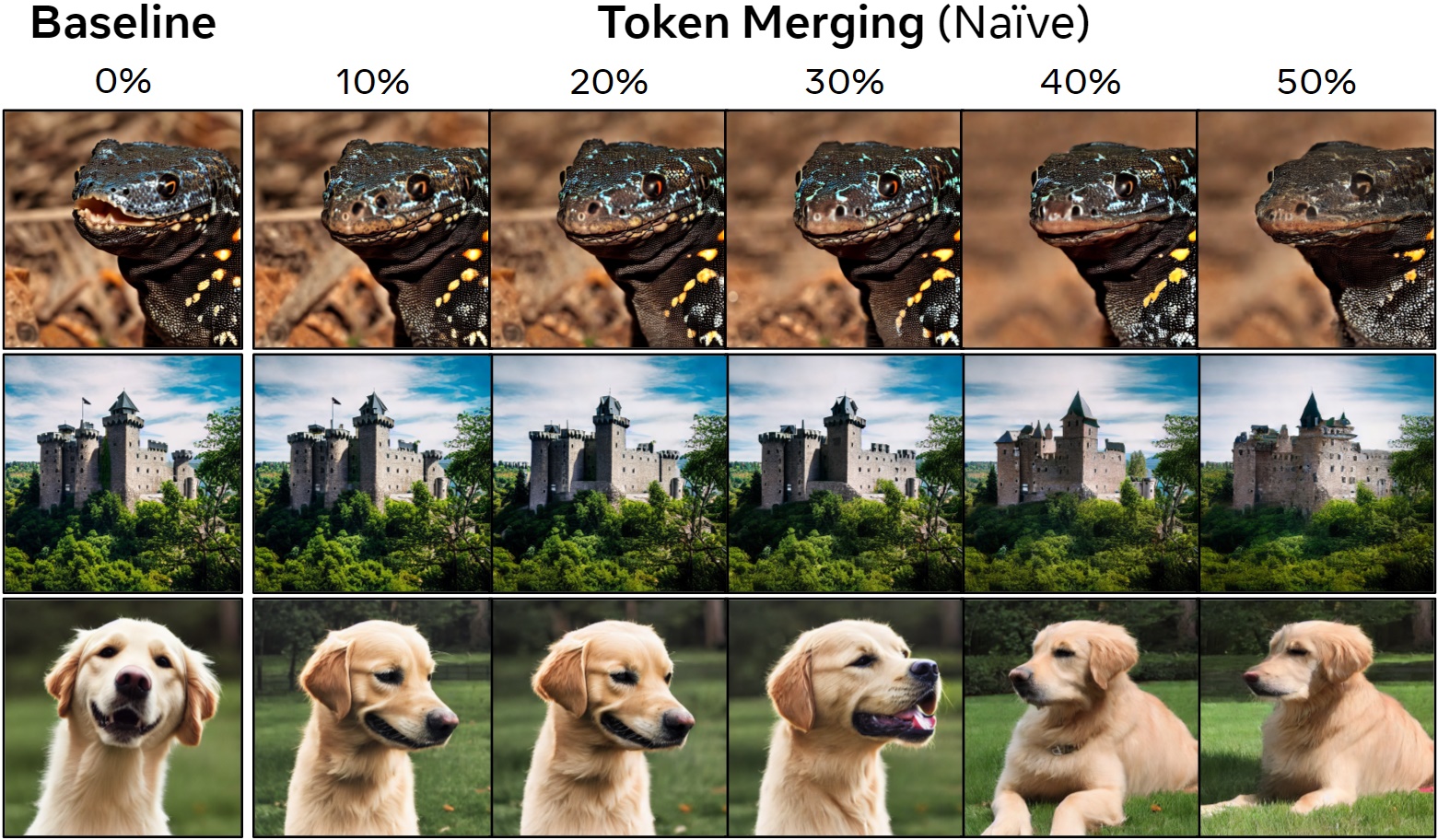}
    \vspace{-1.8em}
    \caption{{\bf ToMe applied na\"ively} as described in Sec.~\ref{sec:approach} works to maintain image coherence, but the content inside the image can change drastically with high amounts of reduction. }
    \label{fig:naive_tome_vis}
\end{figure}

\begin{figure}[t]
    \centering
    \includegraphics[width=1\linewidth]{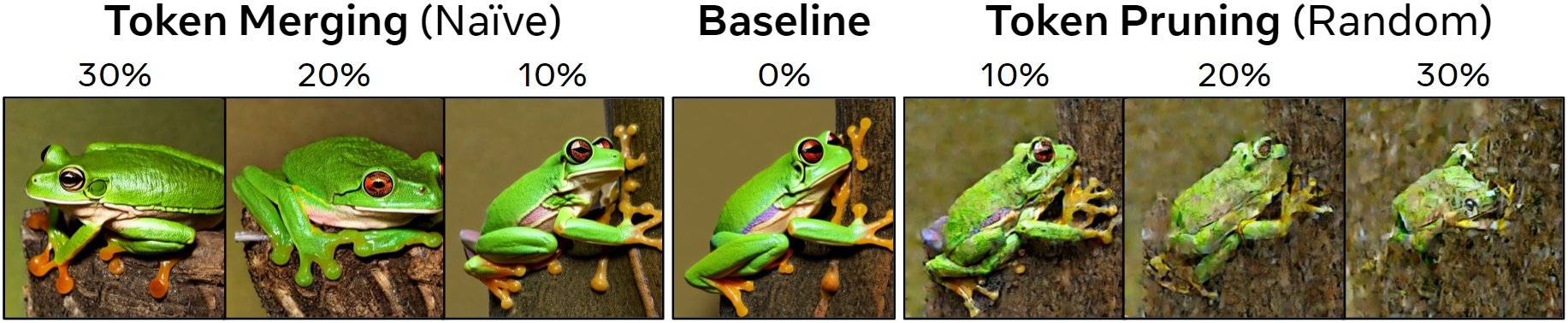}
    \vspace{-1.8em}
    \caption{{\bf Importance of Merging.} If we just prune tokens instead (and replace them with 0), the resulting images quickly degrade. }
    \label{fig:merge_vs_prune}
\end{figure}

In this work, we'll define unmerging in the simplest possible way. Given two tokens with $c$ channels $x_1, x_2 \in \mathbb{R}^{c}$ s.t. $x_1 \approx x_2$, if we merge them into a single token $x^*_{1,2}$, e.g.,
\begin{equation}
    x_{1,2}^* = (x_1 + x_2) / 2
\end{equation}
we can ``unmerge'' them back into $x_1'$ and $x_2'$ by setting 
\begin{equation}
    x_1' = x_{1,2}^* \qquad x_2' = x_{1,2}^*
\end{equation}
While this loses information, the tokens were already similar so the error is small. 
We find this works well in our case, but other unmerging methods might be worth exploring.

\subsection{An Initial Na\"ive Approach}
Merging tokens and then immediately unmerging them doesn't help us though. Instead, we'd like to merge tokens, do some (now reduced) computation, and then unmerge them afterward so we don't lose any tokens. Na\"ively, we can just apply ToMe \textit{before} each component of each block (i.e., self attn, cross attn, mlp), and then unmerge the outputs before adding the skip connection (see Fig.~\ref{fig:unet_block}).

\paragraph{Details.} Because we're not accumulating any token reduction (merged tokens are quickly unmerged), we have to merge a lot more than the original ToMe. Thus instead of removing a \textit{quantity} of tokens $r$, we remove a percentage ($r\%$) of all tokens. Moreover, computing token similarities for merging is expensive, so we only do it once at the start of each block. Finally, we don't use proportional attention and use the input to the block $x$ for similarly rather than attention keys $k$. More exploration is necessary to find if these techniques carry over from the classification setting.

\begin{table}[t]
\centering
    \tablestyle{4pt}{1.05}
    \begin{tabular}{y{50}x{20}|x{30}x{30}x{35}}
        Method & r\% & FID $\downarrow$ & s/im $\downarrow$ & GB/im $\downarrow$ \\
        \shline
        \gc{Baseline} & \gc{0} & \gc{33.12} & \gc{3.09} & \gc{3.41} \\[1pt]
        \textbf{ToMe} (Na\"ive) & 10 & {33.14} & 2.60 & 2.99 \\
         & 20 & 33.53 & 2.29 & 2.17 \\
        & 30 & 33.60 & 2.11 & 1.71 \\
        & 40 & 34.67 & 1.81 & 1.26 \\
        & 50 & 38.95 & {1.53} & 0.89 \\
    \end{tabular}
    \vspace{-1em}
    \caption{{\bf Quantitative evaluation} of the results in Fig.~\ref{fig:naive_tome_vis}. While the approach in Sec.~\ref{sec:approach} can lead to $\sim$2$\times$ faster image generation with $\sim$4$\times$ less memory used (here for $512\times512$ images), it results in a significantly higher FID score. Thus, we explore further. }
    \label{tab:naive_tome}
\end{table}

\section{Further Exploration} \label{sec:experiments}
Amazingly, the simple approach described in Sec.~\ref{sec:approach} works fairly well out of the box \textit{without any training}, even for large amounts of token reduction (see Fig.~\ref{fig:naive_tome_vis}). This is in stark contrast to if we \textit{pruned} tokens instead, which completely destroys the image (see Fig.~\ref{fig:merge_vs_prune}). However, we're not done yet. While the images with ToMe applied look alright, the content within each image changes drastically (mostly for the worse). Thus, we make further improvements using Na\"ive ToMe with 50\% reduction as our starting point.

\paragraph{Experimental Details.} To quantify performance, we use Stable diffusion v1.5 to generate 2,000 512$\times$512 images of ImageNet-1k \cite{imagenet} classes (2 per class) using 50 PLMS \cite{plms} diffusion steps with a cfg scale \cite{guided-diffusion} of 7.5. We then compute FID \cite{fid} scores between those 2,000 samples and 5,000 class-balanced ImageNet-1k val examples using \cite{fid-pytorch}. To test speed, we simply average the time taken over all 2,000 samples on a single 4090 GPU. Applying ToMe na\"ively increases FID substantially (see Tab.~\ref{tab:naive_tome}), though evaluation is up to $2\times$ faster with up to $4\times$ less memory used.

\subsection{A New Partitioning Method} \label{sec:experiments_partitioning_method}
\begin{figure}[t]
    \centering
    \includegraphics[width=1\linewidth]{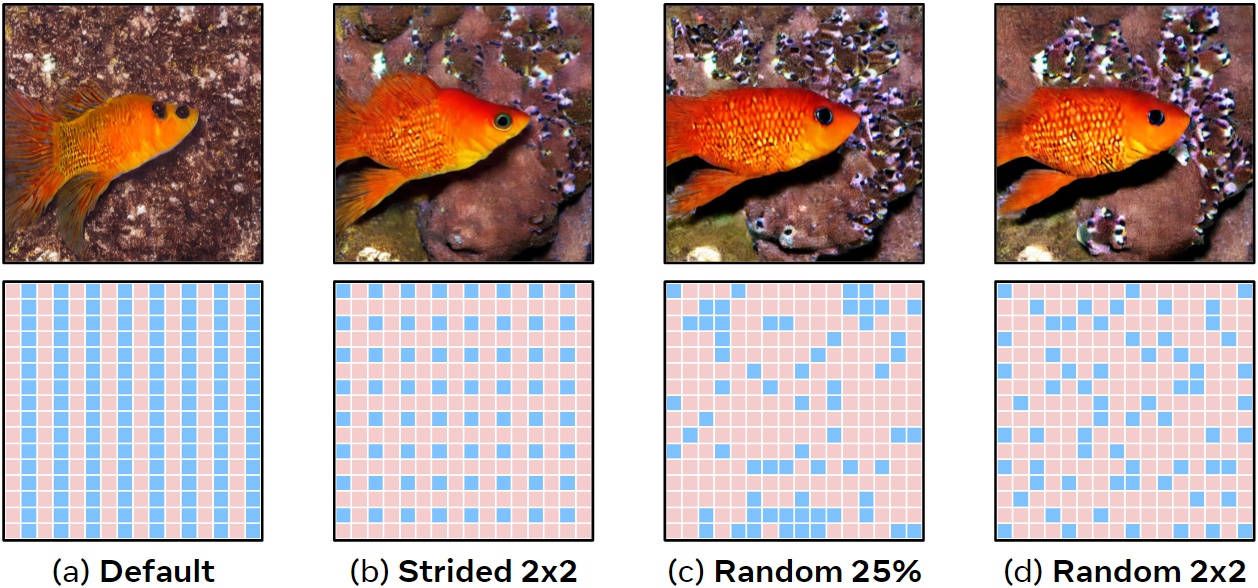}
    \vspace{-1.6em}
    \caption{{\bf Partitioning \src{src} and \dst{dst}.}
    ToMe \cite{tome} merges tokens from \src{src} \textit{into} \dst{dst}.
    (a) By default, ToMe alternates \src{src} and \dst{dst} tokens. In our case, this causes \dst{dst} to form regular columns which leads to bad outputs (poor fish). (b) We can improve generation by sampling \dst{dst} with a 2d stride (e.g., $2\times2$), but this still forms a regular grid. (c) We can introduce irregularity by sampling randomly, but this can cause undesirable clumps of \dst{dst} tokens. (d) Thus, we sample one \dst{dst} token randomly in each $2\times2$ region.  }
    \label{fig:destination_set}
\end{figure}

By default, ToMe partitions the tokens into \src{src} and \dst{dst} (see Sec.~\ref{sec:background}) by alternating between the two. This works for ViTs without unmerging, but in our case this causes \src{src} and \dst{dst} to form alternating columns (see Fig.~\ref{fig:destination_set}\hyperref[fig:destination_set]{a}). Since half of all tokens are in \src{src}, if we merge 50\% of all tokens, then the entirety of \src{src} gets merged into \dst{dst}, so we effectively halve the resolution of the images along the rows.

A simple fix would be to select tokens for \dst{dst} with some 2d stride. This significantly improves the image both qualitatively (Fig.~\ref{fig:destination_set}\hyperref[fig:destination_set]{b}) and quantitatively (Tab.~\ref{tab:partition_ablations_strided}) and gives us the ability to merge more tokens if we want (i.e., the \src{src} set is larger), but the \dst{dst} tokens are still always in the same place. To resolve this, we can introduce randomness.

However, if we just sample \dst{dst} randomly, the FID jumps massively (Tab.~\ref{tab:partition_ablations_random} w/o fix). Crucially, we find that when using classifier-free guidance \cite{guided-diffusion}, the prompted and unprompted samples \textit{need to allocate \dst{dst} tokens in the same way}. We resolve this by fixing the randomness across the batch, which improves results past using a 2d stride (Fig.~\ref{fig:destination_set}\hyperref[fig:destination_set]{c}, Tab.~\ref{tab:partition_ablations_random} w/ fix). Combining the two methods by randomly choosing one \dst{dst} token in each $2\times2$ region performs even better (Fig.~\ref{fig:destination_set}\hyperref[fig:destination_set]{d}), so we make this our default going forward.

\begin{table}[t]
\centering
\subfloat[
    \textbf{Strided} w/ diff strides.
    \label{tab:partition_ablations_strided}
]{
    \centering
    \begin{minipage}{0.45\linewidth}{
        \begin{center}
            \tablestyle{4pt}{1.05}
            \begin{tabular}{x{30}x{20}x{30}}
                $s_y\times s_x$ & \dst{dst}\% & FID $\downarrow$ \\
                \shline
                $1\times2$ & 50 & 38.95 \\
                $2\times1$ & 50 & 39.28 \\
                $2\times2$ & 25 & \default{{\bf 36.12}} \\
                $2\times4$ & 12.5 & 37.09 \\
                $4\times2$ & 12.5 & 37.14 \\
                $4\times4$ & 6.25 & 38.97 \\
            \end{tabular}
    \end{center}}
    \end{minipage}
}
\hfill
\subfloat[
    \textbf{Random} w/ diff methods.
    \label{tab:partition_ablations_random}
]{
    \centering
    \begin{minipage}{0.45\linewidth}{
        \begin{center}
            \includegraphics[width=0.95\linewidth]{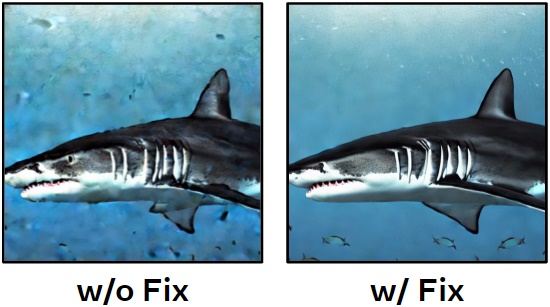}\\
            \vspace{0.2em}
            \tablestyle{4pt}{1.05}
            \begin{tabular}{y{36}x{8}x{26}}
                method & fix & FID $\downarrow$ \\
                \shline
                rand 25\% & \xmark{} & 46.08 \\ 
                rand 25\% & \checkmark{} & 36.00 \\ 
                rand $2\times2$ & \checkmark{} & \default{{\bf 35.66}} \\
            \end{tabular}
    \end{center}}
    \end{minipage}
}
\vspace{-0.4em}
\caption{\textbf{Partition Experiments.} Evaluation of the \src{src} and \dst{dst} partitioning methods described in Fig.~\ref{fig:destination_set}. 50\% of tokens are merged in all experiments. Random methods (b) perform the best provided we fix the randomness across the batch (see Sec.~\ref{sec:experiments_partitioning_method}). }
\label{tab:partition_ablations}
\end{table}

\begin{table*}[t]
\centering
\subfloat[
    \textbf{\textit{What} should we apply ToMe to?} By only applying ToMe to self-attention modules, we can get most of the speed-up with much better FID.
    \label{tab:ablations_what}
]{
    \centering
    \begin{minipage}{0.31\linewidth}{
        \begin{center}
            \tablestyle{4pt}{1.05}
            \begin{tabular}{x{12}x{17}x{12}x{26}x{26}}
                self & cross \\
                attn & attn & mlp & FID $\downarrow$ & s/im $\downarrow$ \\
                \shline
                \checkmark{} & \checkmark{} & \checkmark{} & 35.66 & {\bf1.56} \\
                \checkmark{} & \xmark{} & \checkmark{} & 36.10 & {\bf 1.57} \\
                \checkmark{} & \xmark{} & \xmark{} & \default{{\bf 33.73}} & \default{1.64} \\
                \xmark{} & \xmark{} & \checkmark{} & 34.70 & 2.81 \\
            \end{tabular}
    \end{center}}
    \end{minipage}
}
\hfill
\subfloat[
    \textbf{\textit{Where} should apply ToMe?}
    If we apply ToMe to only the layers with the most tokens, we can get great FID while still being fast.
    \label{tab:ablations_where}
]{
    \centering
    \begin{minipage}{0.31\linewidth}{
        \begin{center}
            \tablestyle{4pt}{1.05}
            \begin{tabular}{x{26}x{32}x{26}x{26}}
                min    &  & \\
                tokens & blocks & FID $\downarrow$ & s/im $\downarrow$ \\
                \shline
                64 & 15 (all) & 35.66 & {\bf 1.56} \\
                256 & 14 & 35.71 & {\bf 1.55} \\
                1024 & 9 & 34.37 & {\bf 1.56} \\
                4096 & 4 & \default{{\bf 33.29}} & \default{1.63} \\
            \end{tabular}
    \end{center}}
    \end{minipage}
}
\hfill
\subfloat[
    \textbf{\textit{When} should we apply ToMe?}
    We can get a small boost by merging more tokens during early diffusion steps and a fewer during later steps.
    \label{tab:ablations_when}
]{
    \centering
    \begin{minipage}{0.31\linewidth}{
        \begin{center}
            \tablestyle{4pt}{1.05}
            \begin{tabular}{x{30}x{26}x{26}x{26}}
                r\% start & r\% end & FID $\downarrow$ & s/im $\downarrow$ \\
                \shline
                70 & 30 & 35.89 & 1.65 \\
                60 & 40 & {\bf 35.53} & 1.58 \\
                50 & 50 & \default{35.66} & \default{\bf 1.56} \\
                40 & 60 & 36.09 & 1.58 \\
                30 & 70 & 36.45 & 1.61 \\
            \end{tabular}
    \end{center}}
    \end{minipage}
}
\vspace{-0.4em}
\caption{\textbf{Design Experiments.} Using the random $2\times2$ partitioning method from Tab.~\ref{tab:partition_ablations}, we now explore how best to apply ToMe (with $r=50\%$). Each experiment is independent, and we highlight our resulting design choice in \colorbox{defaultcolor}{gray}. }
\label{tab:ablations}
\end{table*}

\subsection{Design Experiments}
In Sec.~\ref{sec:approach}, we apply ToMe to every module, layer, and diffusion step. Here we search for a better design (Tab.~\ref{tab:ablations}).

\paragraph{\textit{What} should we apply ToMe to?}
Originally, we applied ToMe to all modules (self attn, cross attn, mlp). In Tab.~\ref{tab:ablations_what}, we test applying ToMe to different combinations of these modules and find that in terms of speed vs.\ FID trade-off, just applying ToMe to self attn is the clear winner. Note that FID doesn't consider prompt adherance, which is likely why merging the cross attn module actually reduces FID.

\paragraph{\textit{Where} should we apply ToMe?}
Applying ToMe to every block in the network is not ideal, since blocks at deeper U-Net scales have much fewer tokens. In Tab.~\ref{tab:ablations_where}, we try restricting ToMe to only blocks with some minimum number of tokens and find that only the blocks with the most tokens need ToMe applied to get most of the speed-up.

\paragraph{\textit{When} should we apply ToMe?}
It might not be right to reduce the same number of tokens in each diffusion step. Earlier diffusion steps are coarser and thus might be more forgiving to errors. In Tab.~\ref{tab:ablations_when}, we test this by linearly interpolating the percent of tokens reduced and find that indeed merging more tokens earlier and fewer tokens later is slightly better, but not enough to be worth it.

\begin{figure}[t]
    \centering
    \includegraphics[width=1\linewidth]{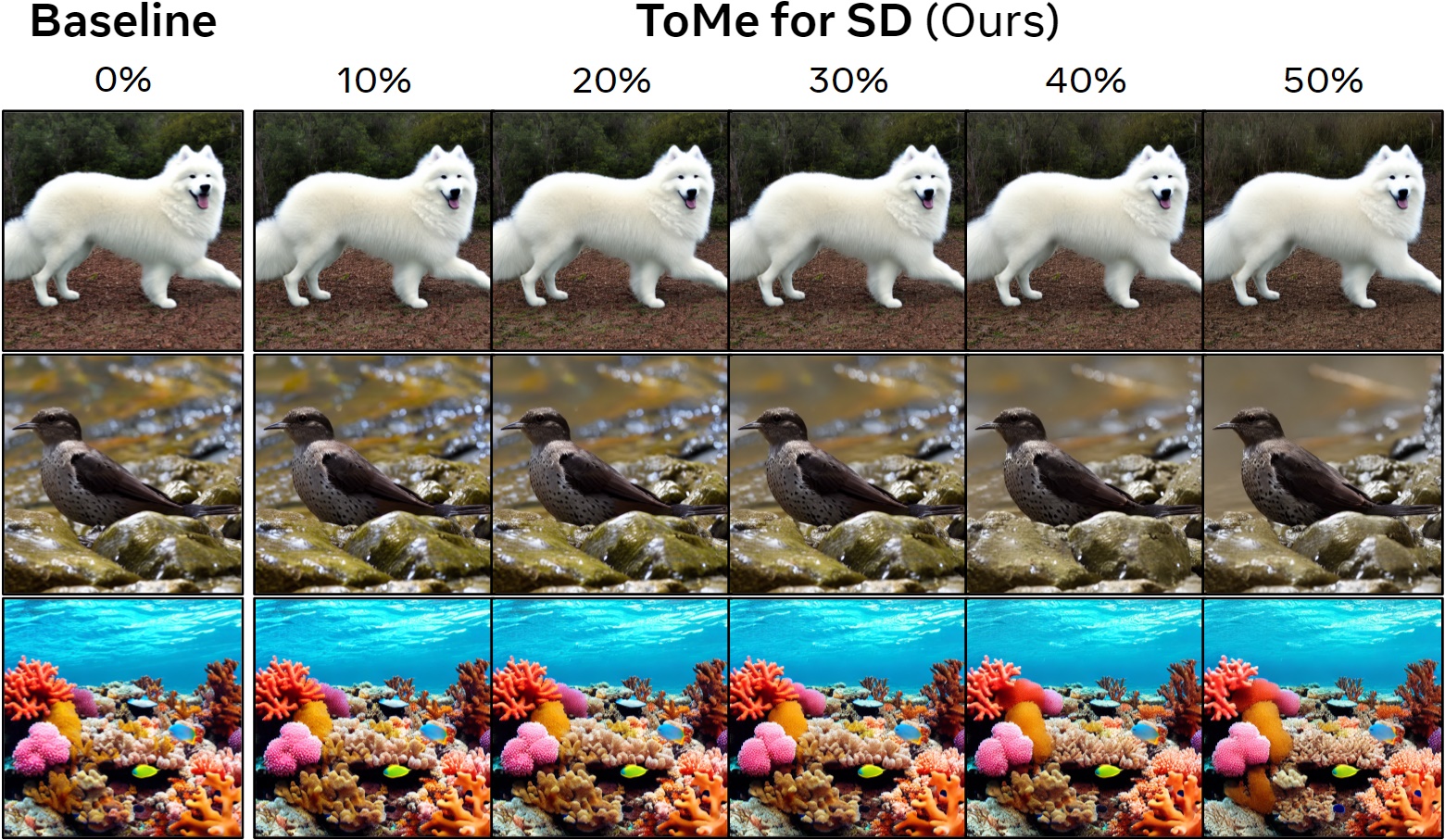}
    \vspace{-1.8em}
    \caption{{\bf Qualitative Results.} Our version of ToMe retains the image content, even at high token reduction. While some slight detail may be lost (e.g., in the background for the dog and bird), our method still handles complex scenes well (like the coral reef). }
    \label{fig:results}
\end{figure}

\begin{table}[t]
\centering
    \tablestyle{4pt}{1.05}
    \begin{tabular}{y{50}x{20}|x{30}x{30}x{35}}
        Method & r\% & FID $\downarrow$ & s/im $\downarrow$ & GB/im $\downarrow$ \\
        \shline
        \gc{Baseline} & \gc{0} & \gc{33.12} & \gc{3.09} & \gc{3.41} \\[1pt]
        \textbf{ToMe for SD} & 10 & 32.86 & 2.56 & 2.99\\
        & 20 & 32.86 & 2.29 & 2.17 \\
        & 30 & 32.80 & 2.06 & 1.71 \\
        & 40 & 32.87 & 1.85 & 1.26 \\
        & 50 & 33.02 & 1.65 & 0.89 \\
        & 60 & 33.37 & 1.52 & 0.60 \\
    \end{tabular}
    \vspace{-1em}
    \caption{{\bf Quantitative Results.} Using the improvements from Sec.~\ref{sec:experiments}, our ToMe for Stable Diffusion obtains similar or better FID compared to the baseline while still being up to $2\times$ faster and using up to $5.6\times$ less memory with 60\% of tokens reduced.  }
    \label{tab:results}
\end{table}

\section{Putting It All Together} \label{sec:results}
We combine all the techniques discussed in Sec.~\ref{sec:experiments} into one method, dubbed ``ToMe for Stable Diffusion''. In Fig.~\ref{fig:results} we show representative samples of how it performs visually, and in Tab.~\ref{tab:results} we show the same qualitatively. Overall, ToMe for Stable Diffusion minimally impacts visual quality while offering up to $2\times$ faster evaluation using 5.6$\times$ less memory.

\begin{figure}[t]
    \centering
    \includegraphics[width=1\linewidth]{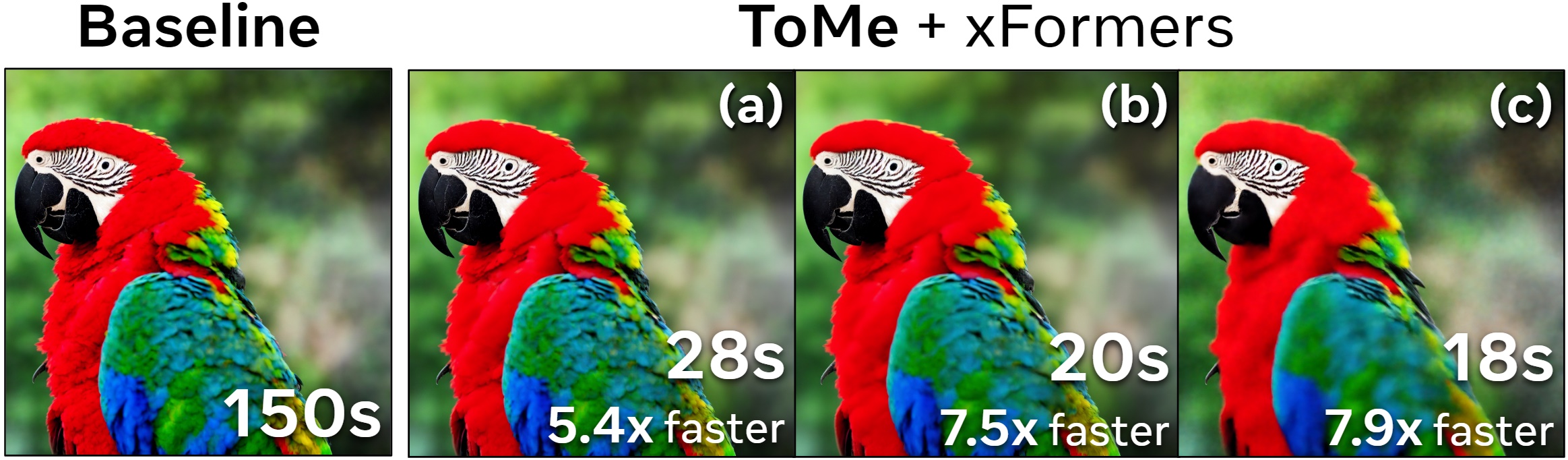}
    \vspace{-1.8em}
    \caption{{\bf Tuning for More Speed-up.} The result in Fig.~\ref{fig:teaser} (a) uses the choices made in Sec.~\ref{sec:experiments}, which are tuned for the best quality. However, we can get even more speed-up if we are okay with some drop in fidelity (b), but there's a limit (c) after which most time is taken by modules outside of our control, resulting in little further speed-up. (Hint: the bird's plumage gets less detailed). }
    \label{fig:more_speedup}
\end{figure}

\paragraph{ToMe + xFormers.}
Since ToMe just reduces the number of tokens, we can still use off the shelf fast transformer implementations to get even more benefit. In Fig.~\ref{fig:teaser} we test generating a $2048\times2048$ image with ToMe and xFormers combined and find massive speed benefits. We can get even more speed-up if we're okay with sacrificing more visual quality (Fig.~\ref{fig:more_speedup}). Note that with smaller images, we found this speed-up to be less pronounced, likely due to the diffusion model not being the bottleneck. Moreover, the memory benefits did not stack with xFormers.

\vspace{-0.15em}
\section{Conclusion and Future Directions}
\vspace{-0.15em}
Overall, we successfully apply ToMe to Stable Diffusion in a way that generates high quality images while being \textit{significantly faster}. Notably, we do this \textit{without training} which is rather remarkable, as any other token reduction method would require retraining. Still, these results could likely be improved by exploring 1.) better unmerging strategies or 2.) whether proportional attention or key-based similarity are useful for diffusion. Furthermore, our success motivates more exploration into using ToMe for dense prediction tasks. We hope this work can serve as both a useful tool for practitioners as well as a starting point for future research in token merging.

\clearpage

\begin{figure*}[h]
    \centering
    \includegraphics[width=0.96\linewidth]{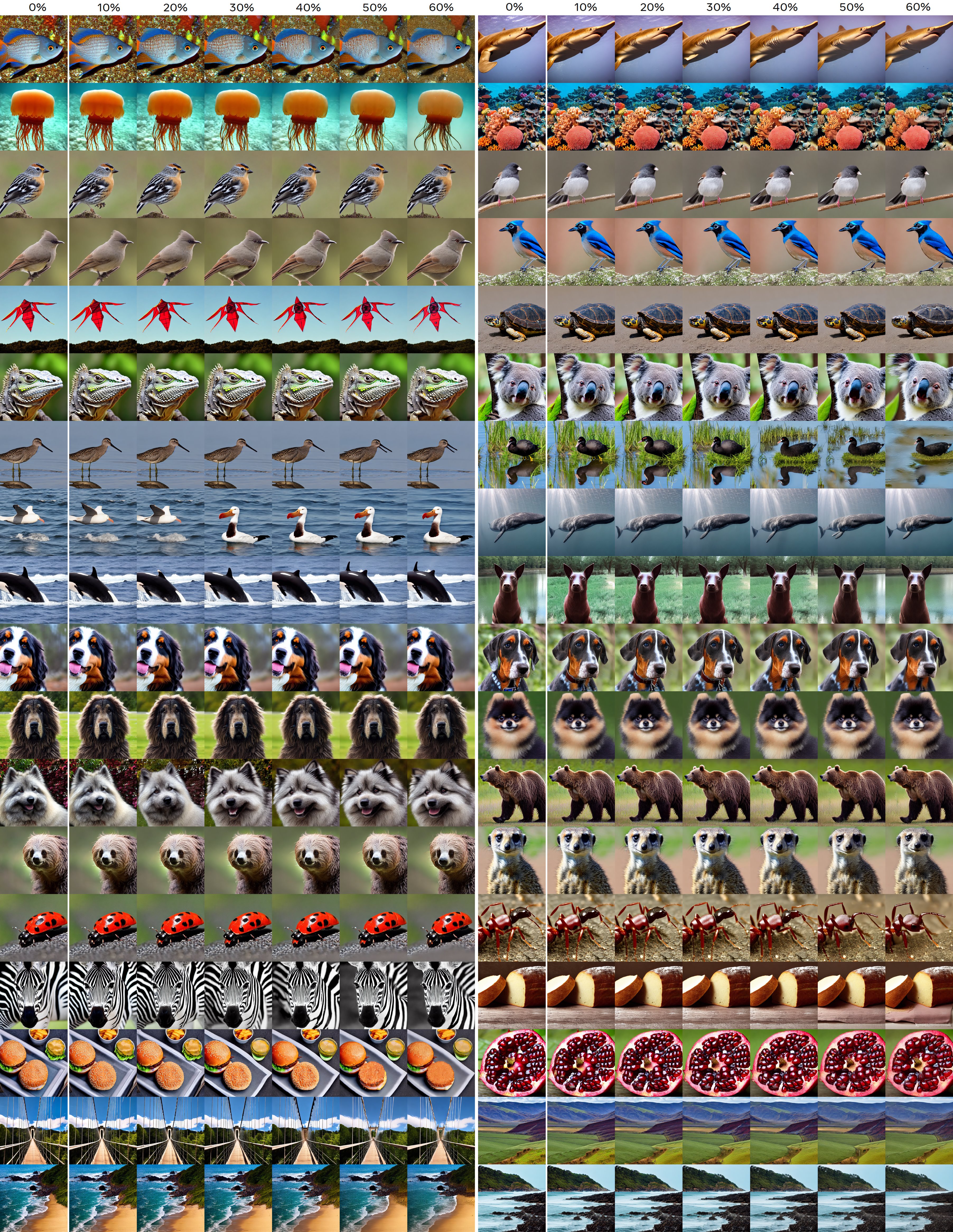}
    \caption{{\bf More examples of ToMe for SD.} Even at 60\% of tokens merged, ToMe for SD keeps the image the same (most of the time).}
    \label{fig:more_examples}
\end{figure*}
\clearpage

{\small
\bibliographystyle{template/ieee_fullname}
\bibliography{main}
}

\end{document}

%% file: definitions.tex
\newlength\savewidth\newcommand\shline{\noalign{\global\savewidth\arrayrulewidth
  \global\arrayrulewidth 0.75pt}\hline\noalign{\global\arrayrulewidth\savewidth}}
\newcommand{\tablestyle}[2]{\setlength{\tabcolsep}{#1}\renewcommand{\arraystretch}{#2}\centering\footnotesize}
\renewcommand{\paragraph}[1]{\vspace{1.25mm}\noindent\textbf{#1}}

\newcolumntype{x}[1]{>{\centering\arraybackslash}p{#1pt}}
\newcolumntype{y}[1]{>{\raggedright\arraybackslash}p{#1pt}}
\newcolumntype{z}[1]{>{\raggedleft\arraybackslash}p{#1pt}}

\newcommand{\app}{\raise.17ex\hbox{$\scriptstyle\sim$}}

\definecolor{deemph}{gray}{0.6}
\newcommand{\gc}[1]{\textcolor{deemph}{#1}}
\definecolor{defaultcolor}{gray}{.9}
\newcommand{\default}[1]{\cellcolor{defaultcolor}{#1}}

\definecolor{dt}{HTML}{ADCAD8}
\definecolor{dt2}{HTML}{cddfe7}

\definecolor{srccolor}{HTML}{bc2f15}
\definecolor{dstcolor}{HTML}{4f83c2}

\newcommand{\src}[1]{\textbf{\textcolor{srccolor}{#1}}}
\newcommand{\dst}[1]{\textbf{\textcolor{dstcolor}{#1}}}

\newcommand{\xmark}{\text{\ding{55}}}%